\documentclass[runningheads]{llncs}
\usepackage[T1]{fontenc}
\usepackage{graphicx}
\usepackage{amsmath}
\usepackage{mathtools}
\begin{document}
\title{Driving down Poisson error can offset classification error in clinical tasks}
%
%
\author{Charles B. Delahunt\inst{1*}  \and
Courosh Mehanian\inst{1} \and
Matthew P. Horning\inst{1} }
\authorrunning{Delahunt et al.}
%
\institute{\textsuperscript{1} Global Health Labs, Bellevue, WA \\
\email{\textsuperscript{*}charles.delahunt@ghlabs.org}}
%
\maketitle              
\begin{abstract}

Medical machine learning algorithms are typically evaluated based on accuracy vs. a clinician-defined ground truth, a reasonable initial choice since trained clinicians are usually better classifiers than ML models. 
However, this metric does not fully capture the actual clinical task: it neglects the fact that humans, even with perfect accuracy, are subject to non-trivial error from the Poisson statistics of rare events, because clinical protocols often specify a relatively small sample size. 
For example, to quantitate malaria on a thin blood film a clinician examines only 2000 red blood cells (0.0004 $\mu L$), which can yield large Poisson variation in the actual number of parasites present, so that a perfect human's count can differ substantially from the true average load. 
In contrast, an ML system may be less accurate on an object level, but it may also have the option to examine more blood (e.g. 0.1 $\mu L$, or 250$\times$). 
Then while its parasite identification error is higher, the Poisson variability of its estimate is lower due to larger sample size.

To qualify for clinical deployment, an ML system's performance must match current standard of care, typically a very demanding target. 
To achieve this, it may be possible to offset the ML system's lower accuracy by increasing its sample size to reduce Poisson error, and thus attain the same net clinical performance as a perfectly accurate human limited by smaller sample size. 
In this paper, we analyse the mathematics of the relationship between Poisson error, classification error, and total error.
This mathematical toolkit enables teams optimizing ML systems to leverage a relative strength (larger sample sizes) to offset a relative weakness (classification accuracy). We illustrate the methods with two concrete examples: diagnosis and quantitation of malaria on blood films.
  
\end{abstract} 
\section{Introduction}
For ML solutions to be clinically useful, their performance must match or exceed current clinical requirements. 
During proof-of-concept, we typically evaluate ML models against a human defined ground truth, and judge its performance against human accuracy. 
However, this is not the only relevant axis of performance at the clinical task. Humans and ML have different strengths and weaknesses, and ML's strengths can be leveraged to offset its weaknesses to meet a clinical performance goal.
Humans are highly adaptable and accurate at tasks like identifying malaria parasites, but they examine only a limited sample size due to time and fatigue constraints.
This sample size is typically encoded in clinical protocols such as how much blood to examine for malaria quantitation, or how many white blood cells (WBCs) to inventory for a differential blood count. 
In contrast, machines often have lower accuracy at tasks such as identifying malaria parasites, but they don’t fatigue and can potentially examine a much larger sample. 

In this paper we examine the relationship between Poisson error (variation in counts of rare objects in a small sample), accuracy error (object classification error), and total error (deviation from the expected value in a very large sample).
Limited sample sizes encoded in clinical protocols mean that humans, even with perfect classification accuracy, are subject to non-trivial Poisson error relative to the background ground truth (e.g. average parasites per $\mu L$ of blood).
We study how an automated system can offset its imperfect classifier accuracy by examining a larger sample and thus reducing its Poisson error, resulting in equal net error relative to a human following a clinical protocol. 
That is, machines can use a particular strength (high data throughput) to offset a weakness (classification inaccuracy) and thus match the performance on a clinical task of a human with high classification accuracy but lower data throughput.

We assume the clinicians have perfect (100\%) accuracy\footnote{Not all field microscopy has perfect accuracy; but well-trained and equipped microscopists are very skilled, and any ML system will typically need to match the ``best case'' human performance.}.
However, they are still subject to unavoidable Poisson error when a clinical protocol specifies that they examine a relatively small amount of substrate, e.g. 200 WBCs for a differential blood count, or 0.0625 $\mu L$ of blood for malaria diagnosis. 
We describe how to calculate what increase in examined substrate volume  is required to offset imperfect ML accuracy, such that the ML system does not exceed the total error allowed by current standards of care. 

We address two widely-relevant use cases: (i) limits of detection, (ii) quantitation accuracy. 
We illustrate the calculation methods on specific clinical examples of these use cases, namely (i) diagnosis of malaria and (ii) quantitation of malaria parasites, on blood films.
We provide equations and step-by-step methods to guide how to modify an operational parameter of an ML system (i.e. ``sample size examined'') to offset imperfect object classification accuracy. 
Our mathematical derivations necessarily assume certain (well-principled) starting formulas for limit of detection (LoD) and quantitation.
Thus if an algorithm uses very different starting formulas then different mathematical derivations / calculations will be required. 
In this case, our paper serves to illustrate how one can go about analyzing the relationship between Poisson error, classification error, and total error, as a toolkit to leverage the benefits of larger sample size.

Our examples highlight the methods' relevance to parasitic diseases, and they have clear application in low resource regions where microscopy is a central tool. 
But the methods have utility whenever sample sizes specified by protocols imply non-trivial Poisson variability.  ``Parasite'' can be replaced with ``object of interest'' (e.g. abnormal cell).
This paper is relevant to ML systems that move beyond the academic proof-of-concept and aspire to deployment in the clinic.  
 
The next two sections describe mathematics for (i) diagnosis and LoD, and (ii) quantitation, in each case illustrated with concrete medical use cases in which some of the variables in the equations are known (e.g. fixed by protocols). 
\section{Diagnosis and limit of detection}

Diagnosis of malaria \cite{whoMicroscopyLearners2010,whoQualityAssuranceV2} and neglected tropical diseases (e.g. schistosomiasis, lymphatic filariasis) \cite{whoHelminths2002,whoHelminths2021} at the low parasitemias near LoD is a case of rare object detection, in which an examined sample might contain only a few parasites and the exact number present in the sample is subject to Poisson variability:
\begin{equation}
P(k~events) = \frac{\lambda^k e^{-\lambda}}{k!}
\label{eqnPoisson}
\end{equation}
where $k$ is an integer and $\lambda$ is the expected number of events in a given interval.

In our case, ``events'' are parasites, ``interval'' is the volume of substrate (e.g. blood) examined, and the ``expected number'' is the parasitemia (e.g. parasites per $\mu L$) scaled by the examined volume. 
For large volumes and/or high parasitemias, this basically matches the binomial distribution, and diagnosis is not an issue (it does affect quantitation, as described in Section \ref{secQuantitation}). 
At small volumes and low parasitemias, the Poisson distribution squeezes up against the y-axis, giving a larger standard error (i.e. std dev / mean = $\sigma (k) / \mu (k)$), and giving an asymmetrical distribution in which $k = 0$ has nontrivial probability.
Python \cite{van1995python} code to plot Poisson distributions, used for computational sweeps, is given in Appendix \ref{appendixCode}.
This low parasitemia regime is relevant to diagnosis at LoD.
\subsection{Example: Malaria}
\label{subsecMalariaLoD}
We assume that a human can perfectly classify objects in blood as a malaria parasite or artifact. 
Then, since there are no false positive artifacts to contend with, the LoD = $N$ p/$\mu L$ is the parasitemia at which the examined sample volume consistently (e.g. 95\% of the time) contains at least one parasite .

Let $n$ = number of parasites in the examined volume $V$; $cV$ = clinically-relevant volume ($1 ~\mu L$ for malaria), and $N/cV$ be the parasitemia at LoD. Then we require that
\begin{equation}
P(n \geq 1 ~| ~V, ~N) \geq 0.95
\label{eqnLoD}
\end{equation}
where the underlying distribution is Poisson. 

\begin{figure}
\begin{center}
\includegraphics[width=0.6\textwidth]{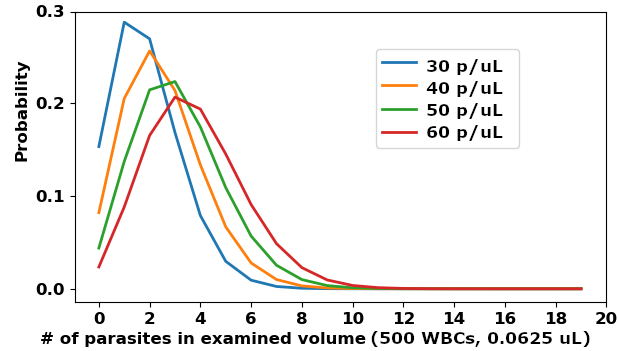}
\caption{Poisson distributions showing number of parasites actually found in 500 WBCs' worth of blood, given various true parasitemias. The LoD is $\approx$50 p/$\mu L$.} 
\label{figPoisson}
\end{center}
\end{figure}

The WHO guidelines for malaria microscopy \cite{whoMicroscopy2016} specify that $V$ contain 500 WBCs, $\approx$ 0.0625 (i.e.  $1/16^{th}$) $\mu L$ using the standard approximation of 8000 WBCs/$\mu L$. 
Plugging in values for $N$, we find an LoD of $\approx$50 p/$\mu L$ (see Figure~\ref{figPoisson}).
We assume that our ML system has imperfect object-level accuracy, and follow the analysis in \cite{frontierMetrics}.
For a given patient, let object sensitivity = $s$ and false positive rate $f$ = FPs/$cV$. 
Then let \textit{\textbf{S}} be the vector of object-level sensitivities $s$ over each patient and \textit{\textbf{F}} be the vector of FP rates $f$ over each patient.
We note that $f$ varies from patient to patient and consider $\sigma(\textit{\textbf{F}})$, the standard deviation of FP rates over the population of patients.
For simplicity we neglect the variation in $s$ between patients and use the mean $\mu (\textit{\textbf{S}})$ (for more nuance on this, see \cite{frontierMetrics}).

We wish to calculate how large an examined volume $V$ the ML system needs to examine, such that it will reliably (e.g. 95\% of the time) both identify malaria-negative patients and also detect malaria-positive patients at the LoD of $N$ p/$\mu L$. 
To achieve high patient-level specificity, we set a threshold $T$ on the number of suspected parasites detected by the model, such that for most negative patients the suspected parasite count will come in below threshold. Assuming a Gaussian distribution, we can define $T$ as:
\begin{equation}
T = \Bigl( \mu ({\bf{F}}) + 1.65 \sigma ({\bf{F}})\Bigr)~ \frac{V}{1~\mu L}
\label{eqnThresholdForSpec}
\end{equation}
where the scaling term gives us the number of FPs in volume $V$. 
Then to achieve high sensitivity on samples with parasitemia at the LoD $N$ p/$\mu L$ we need the the number of suspected parasites (the sum of true positives TP and false positives FP) to equal or exceed this threshold.
This must hold for most patients. 
So it must hold for a patient with a relatively ``clean'' sample, i.e.  with very few FPs (the bottom end of the FP rate distribution):
\begin{equation}
\#FPs = \Bigl( \mu ({\bf{F}}) - 1.65 \sigma ({\bf{F}}) \Bigr)~ \frac{V}{1~\mu L}
\label{eqnClean}
\end{equation}
Since the suspected parasite count in $V$ equals 
$\# TP ~ + ~ \# FP $, we need 
\begin{equation}
 \# TP > 3.3 \sigma ({\bf{F}})~\frac{V}{1~\mu L} 
 \label{eqnTpformula}
\end{equation}  
Given sensitivity $\mu (\textit{\textbf{S}})$, we need at least $\#TP / \mu (\textit{\textbf{S}})$ true parasites present in the examined volume  to find sufficient TPs to cross the threshold $T$ (as required by Equation~\ref{eqnTpformula}). 
Note that the number of TPs we need to have present in $V$ depends on $V$, because as $V$ increases so does the FP count spread and thus the threshold $T$.
Let $x$ = this required number of TPs.

To find the examined volume $V$ needed to meet this spec, we can computationally sweep values of $V$ and plug $\{V, x(V), N\}$ into the Poisson distribution to see which $V$ gives $P(k \leq x) < 0.05$ (see Python code in Appendix~\ref{appendixCode}). 

\begin{figure}
\begin{center}
\includegraphics[width=0.75\textwidth]{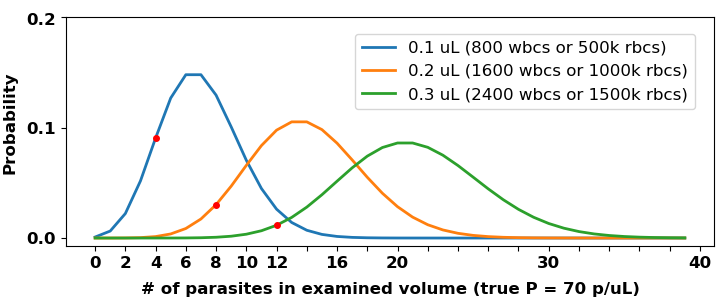}
\caption{Estimating the volume required to match an LoD = 70, given $\mu (\textit{\textbf{S}})$ = 0.85 and $\sigma (\textit{\textbf{F}}) = 10 / \mu L$. The required number of TPs for each volume are shown as red dots. For $V = 0.2~ \mu L$, 95\% of cases will exceed the required number.} 
\label{figPoissonForDifferentVs}
\end{center}
\end{figure}

For example, if our model has $\mu ({\bf{S}})$ = 0.85 and $\sigma ({\bf{F}}) = 10 / \mu L$, then attaining LoD = 50 p/$\mu L$ is not feasible. However, an LoD = 70 p/$\mu L$ can be attained by examining $V = 0.2~ \mu L$ (see Figure~\ref{figPoissonForDifferentVs}). 
This volume is larger than specified by WHO protocol, but is potentially tractable for automated hardware.
\section{Quantitation}
\label{secQuantitation}
Quantitation is sometimes a clinically relevant task. 
For example, to monitor drug-resistant malaria strains, sentinel sites dose \textit{P. falciparum}-positive patients with a drug, then create and quantitate blood films every 4 to 6 hours, to see if the strain is developing resistance to the drug \cite{ashley,white,whoMalariaResearch2016}.
This use case involves a huge work burden for microscopists and has relaxed runtime requirements. 
It is thus an ideal target for automated systems.

Another example concerns mass drug administration of ivermectin to control onchocerciasis.
In regions where the filaria \textit{Loa loa} is co-endemic, patients with high \textit{Loa loa} microfilaria burdens (25,000 - 30,000 mf/\textit{mL} of blood) risk catastrophic side effects from ivermectin. 
Thus accurate quantitation at the 25,000 mf/\textit{mL}  cutoff is clinically vital for the ``Test and Not Treat'' protocol \cite{gardon,dambrosio}.

Poisson error is introduced into quantitation, even at high parasitemias, when a relatively small sample is examined.
Let $\mathcal{P}(p, V)$ = the Poisson distribution for parasitemia $p$ p/$cV$ and examined volume $V$.
Then the standard error caused by Poisson variability at true parasitemia $p$ is \\
\begin{equation}
\frac{\sigma (\mathcal{P}(p, V))}{\mu(\mathcal{P}(p, V))} = \frac{\sigma (\mathcal{P}(p, V))}{ p \frac{V}{cV} } = 
\sqrt{\frac{1}{p}\frac{cV}{V}}
\label{eqnStdErrorForPoisson}
\end{equation}
since for a Poisson distribution the variance equals the mean.
The denominator is the expected number of parasites in $V$ given parasitemia $p$.
This unavoidable error is remarkably high for the small $V_{PR}$ specified by some protocols.

\subsection{Example: quantitation of malaria parasites}
\begin{figure}
\begin{center}
\includegraphics[width=0.8\textwidth]{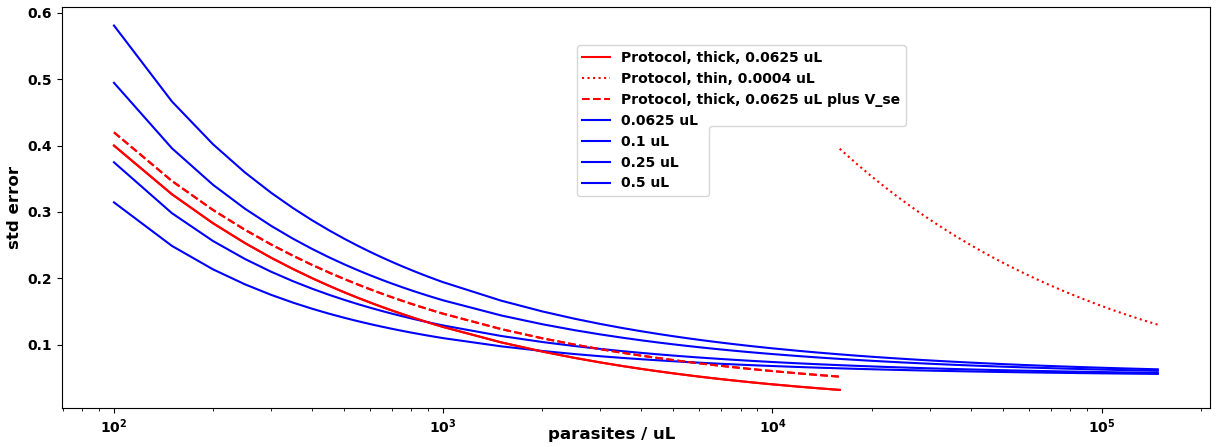}
\caption{Standard error of quantitation vs. parasitemia, for: 
(In RED) Humans with perfect accuracy, one without and one with volume estimation error, examining protocol volumes;
(In BLUE) An imperfect algorithm with performance per the text, at various examined volumes $V$. 
Larger $V$ compensates for its classification and volume errors. }
\label{figPoissonErrorForQuant}
\end{center}
\end{figure}

To quantitate blood films for malaria, WHO research protocols specify examining 500 WBCS (0.0625~ $\mu L$) if $p < 16,000 p/\mu L$, or 2000 red blood cells (RBCs), ($\approx 0.0004 ~\mu L$) if $p > 16,000 p/\mu L$ \cite{whoMalariaResearch2016}.
The unavoidable standard error due to Poisson variability is plotted as red lines in Figure~\ref{figPoissonErrorForQuant}.

We wish to compare the errors in the algorithm quantitation to this Poisson error on $V_{PR}$.
Following \cite{mehanian}, we define our formula for estimated parasitemia in a particular sample as 
\begin{equation}
\hat{p} = \frac{(tp + fp) - \mu(\textit{\textbf{F}}) \frac{V_E}{cV}}{\mu(\textit{\textbf{S}})} ~ \frac{cV}{V_E} 
\label{eqnQuantFormula}
\end{equation}
\noindent where $tp + fp$ = the number of suspect parasites in examined volume $V$ (both true parasites and misclassified distractors),  
$V_E$ is the estimate (e.g. found by counting WBCs or RBCs) of $V$, 
and other terms are as defined previously. 
We let $\textit{\textbf{V}}_E = $ the vector of $V_E$ over the population, in the useful case (below) that all examined volumes are the same $V$.

This formula says: Count up the suspects; then subtract the expected number of FPs in $V_E$, to get an estimate of detected parasites $tp$; divide this by our expected sensitivity ${\mu(\textit{\textbf{S}})}$ to get an estimate of the actual number of parasites that were present; then normalize by estimated volume to get $\hat{p} / cV$.

This estimate contains three sources of error relative to true parasitemia $p$: classification (of both parasites and distractors), volume estimation, and Poisson variability.
These all vary over the patient population. 
We seek a formula for the standard deviation of the combined error $sigma(\textit{\textbf{p}}_E)$ from these three sources.

Let the parasitemia = $p / \mu L$, and $p_V = $ the actual number of parasites in $V$, so $p_V$ has a Poisson distribution with mean $p \frac{V}{cV}$.\\
In what follows, $\Delta $ denotes the difference of the sample's variable from the mean population value, i.e.
$S = \mu(\textit{\textbf{S}}) + \Delta S, ~F = \mu(\textit{\textbf{F}}) + \Delta S, ~V_E = V + \Delta V$, and\\
$p_V = p \frac{V}{cV} + \Delta \mathcal{P}(p,V)$ where $\mathcal{P}(p,V)$ denotes Poisson variability.  
So
\begin{equation} 
tp = p_V~ (\mu(\textit{\textbf{S}}) + \Delta S) = ( p\frac{V}{cV} + \Delta \mathcal{P}(p,V)  )(\mu(\textit{\textbf{S}}) + \Delta S ) 
\end{equation}
\begin{equation}
fp = (\mu(\textit{\textbf{F}}) + \Delta F) \frac{V}{cV} ~~~~~~~~~~~~~~~~~~~~~~~~~~~~~~~~~~~~~~~~~~~~ 
\end{equation}
To get a formula for the standard error of quantitation $\frac{\sigma(\textit{\textbf{p}}_E)}{p}$, we substitute these terms into Equation~\ref{eqnQuantFormula}, apply some algebra, convert into standard deviations over the population, then divide by $p$. 
The full derivation is in Appendix~\ref{appendixDeriveQuantError}:
\begin{equation}
\begin{multlined}
\frac{\sigma(\textit{\textbf{p}}_E)}{p} = 
V_{SE}   ~+~ 
\frac{\sigma(\textit{\textbf{S}})}{\mu(\textit{\textbf{S}})} \left(1 + V_{SE}\right)  ~+~ ~~~~~~~~~~~~~~~~~~~\\
~~~~~~~~~\left(1 + \frac{\sigma(\textit{\textbf{S}})}{\mu(\textit{\textbf{S}})} \right)
 \sqrt{\frac{1}{p}\frac{cV}{V}}  ~+~  
\frac{V_{SE}}{p}\frac{\mu(\textit{\textbf{F}})}{\mu(\textit{\textbf{S}})}   ~+~
\frac{\sigma(\textit{\textbf{F}})}{\mu(\textit{\textbf{S}})} \frac{(1 + V_{SE})}{p}
\end{multlined}
\label{eqnStdQuantError}
\end{equation}
 where $V_{SE} = \frac{\sigma(\textit{\textbf{V}}_E)}{V}$ is a constant (see Appendix~\ref{appendixDeriveQuantError} for details).
 
The crucial thing to note about Equation~\ref{eqnStdQuantError} is 
that the values on the RHS are known:  
$\mu(\textit{\textbf{S}}), \sigma(\textit{\textbf{S}}), \mu(\textit{\textbf{F}}), \sigma(\textit{\textbf{F}}), \text{ and } V_{SE}$ are readily-calculated performance statistics of the algorithm (for full details see \cite{frontierMetrics}).
So Equation~\ref{eqnStdQuantError} is a simple function of $\{p, V\}$.

\begin{figure}
\begin{center}
\includegraphics[width=0.95\textwidth]{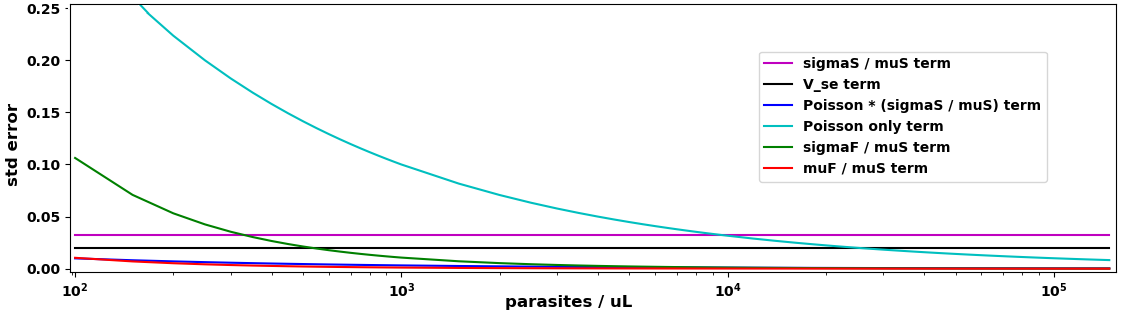}
\caption{The size of various terms in the standard error of quantitation, Equation~\ref{eqnStdQuantError}.
At low $p$ errors due to Poisson and $\sigma (\textit{\textbf{F}})$ dominate, while at high $p$ errors due to $\sigma (\textit{\textbf{S}})$ and $V_{SE}$ dominate. See Python code in Appendix~\ref{appendixStdErrorQuantCode}.} 
\label{figComponentsOfError}
\end{center}
\end{figure}

The contributions of the various terms in Equation~\ref{eqnStdQuantError} are plotted in Figure~\ref{figComponentsOfError}.
For most parasitemias $p$ the biggest non-Poisson contributions are from $\frac{\sigma(\textit{\textbf{F}})}{\mu(\textit{\textbf{S}})}$ and $V_{SE}$. The terms with $\sigma(\textit{\textbf{F}})$ only matter at low $p$.
This plot is a useful tool to highlight where algorithm improvements are most needed.
Our original question was: How much extra volume must we examine to reduce Poisson error enough to offset algorithm errors, such that the automated system's total error matches the Poisson error of a perfect clinician examining a protocol volume $V_{PR}$?

We can answer this by first plotting the curve $\frac{1}{p}\sigma(\mathcal{P}(p,V_{PR}))$ over values of $p$, then plotting curves $\frac{1}{p}\sigma(\textit{\textbf{p}}_E)(p,V)$ from Equation~\ref{eqnStdQuantError} for a set of volumes $V$. 
This is illustrated in Figure~\ref{figPoissonErrorForQuant} for a (strong) hypothetical algorithm with $\mu(\textit{\textbf{S}}) = 0.95, \sigma(\textit{\textbf{S}}) = 0.03, \mu(\textit{\textbf{F}}) = 50, \sigma(\textit{\textbf{F}}) = 10, \text{ and } V_{SE} = 0.02$. 
We find that when applied to $V_{PR}$ the algorithm has far higher standard error than a perfect human following protocol, but examining 0.4 $\mu L$ suffices to closely match the human's error at most parasitemias, especially if, as is likely, the human volume estimate also has error (see \ref{secEstimateVolumeApp} for details).
%
%
 A larger examined blood volume can be attained in two ways: (i) by examining more RBCs on thin film (as in \cite{noul}); or (ii) by staying on thick films (as in \cite{mehanian,das,reesChanner}) since machines do not need to switch to thin film at 16,000 p/$\mu L$, giving the large advantage shown in Figure~\ref{figPoissonErrorForQuant}.
 \section{Discussion}
To deploy into clinical settings, an ML system's performance must match current standards of care, typically a challenging requirement. 
In this paper we have described how standard protocols for human microscopy assessment specify, by necessity, relatively small sample sizes and thus have high levels of unavoidable Poisson variability.
We have also described mathematical methods to analyze the relationship between Poisson error, algorithm error, and total error.
These methods can inform the principled use of increased sample size to offset algorithm error and thus achieve performance equivalent to a perfectly accurate human on a clinical task.

We note that the ability to examine a larger sample is subject to technical, cost, and use case constraints. 
For example, in the drug resistance monitoring use case, time-to-result is relaxed and allows more scanning time; while point-of-care settings require fast diagnosis, limiting scanning time and thus examined volume.
Cost can also constrain increases to sample size.
For example, \cite{noul} uses a cartridge, so increasing examined volume beyond one cartridge's worth doubles the cost of consumables; while a system scanning thick films can examine extra volume at no extra cost.

The ability to drive down Poisson error offers teams that are optimizing automated ML systems for deployment a separate, valuable axis for improvement, an axis which humans cannot realistically leverage.
Improvement on this axis, involving both hardware and software, can offset lower ML performance on tasks at which humans excel, such as object classification detached from the constraints of clinical protocols, enabling ML systems to meet the rigorous clinical performance standards of care required for deployment.


 
%
%
 

\begin{credits}
\subsubsection{\ackname} Funded by Global Health Labs, Inc (\url{www.ghlabs.org}).

\subsubsection{\discintname}
The authors have no competing interests to declare that are
relevant to the content of this article.
\end{credits}

 
\bibliographystyle{splncs04}
\bibliography{bibliography} 

\begin{thebibliography}{10}
\providecommand{\url}[1]{\texttt{#1}}
\providecommand{\urlprefix}{URL }
\providecommand{\doi}[1]{https://doi.org/#1}

\bibitem{ashley}
Ashley, E., Dhorda, M., Fairhurst, R., Amaratunga, C., et~al.: Spread of artemisinin resistance in \textit{{P}lasmodium falciparum} malaria. New England J of Medicine  (2014)

\bibitem{das}
Das, D., Vongpromed, R., Dhorda, M., et~al.: Field evaluation of the diagnostic performance of {EasyScan GO}: a digital malaria microscopy device based on machine-learning. Malaria J  (2022)

\bibitem{thinGhtc2019SuppInfo}
{Delahunt}, C.B., {Jaiswal}, M.S., {Horning}, M.P., {Janko}, S., {Thompson}, C.M., {Kulhare}, S., et~al.: Fully-automated patient-level malaria assessment on field-prepared thin blood film microscopy images, with supplemental information. IEEE GHTC Proceedings, arXiv version  (2019), \url{https://arxiv.org/abs/1908.01901}

\bibitem{frontierMetrics}
Delahunt, C., Gachuhi, N., Horning, M.: Metrics to guide development of machine learning algorithms for malaria diagnosis. in review  (2024), \url{https://arxiv.org/abs/2209.06947}

\bibitem{dambrosio}
D’Ambrosio, M.V., Fletcher, D., et~al.: Point-of-care quantification of blood-borne filarial parasites with a mobile phone microscope. Science Trans Medicine  (2015)

\bibitem{gardon}
Gardon, J., Boussinesq, M., et~al.: Serious reactions after mass treatment of onchocerciasis with ivermectin in an area endemic for loa loa infection. Lancet  (1997)

\bibitem{mehanian}
Mehanian, C., {Horning}, M., et~al.: Computer-automated malaria diagnosis and quantitation using convolutional neural networks. ICCV  (2017)

\bibitem{peruMalaria}
{Ministerio de Salud}: {Manual de Procedimientos de Laboratoria Para el Diagnostico de Malaria} (2003), {Lima, Peru}

\bibitem{noul}
Noul: {miLab} platform  ({2023}), {S. Korea.} \url{https://noul.kr/en/milab-platform/}

\bibitem{reesChanner}
Rees-Channer, R., Bachman, C., Chiodini, P., et~al: Evaluation of an automated microscope using machine learning for the detection of malaria in travelers returned to the {UK}. Frontiers Malaria  (2023)

\bibitem{van1995python}
Van~Rossum, G., Drake~Jr, F.L.: Python reference manual. Centrum voor Wiskunde en Informatica Amsterdam (1995)

\bibitem{white}
White, N.: The parasite clearance curve. Malaria J  (2011)

\bibitem{whoHelminths2002}
{WHO}: Prevention and control of schistosomiasis and soil-transmitted helminthiasis (2002), {World Health Organization, Geneva, Switzerland}

\bibitem{whoMicroscopyLearners2010}
{WHO}: Basic malaria microscopy. Part I. Learner’s guide. 2nd ed, (esp units 7, 8, and 9) (2010), {World Health Organization, Geneva, Switzerland}

\bibitem{whoQualityAssuranceV2}
{WHO}: Malaria microscopy quality assurance manual v2  (2016), {World Health Organization, Geneva, Switzerland}

\bibitem{whoMicroscopyQuantSOP}
{WHO}: Malaria Microscopy Standard Operating Procedure MM-SOP-09: Malaria Parasite Counting (2016), {World Health Organization, Geneva, Switzerland}

\bibitem{whoMicroscopy2016}
{WHO}: Microscopy examination of thick and thin blood films for identification of malaria parasites (esp SOPs 8 and 9) (2016), {World Health Organization, Geneva, Switzerland}

\bibitem{whoMalariaResearch2016}
{WHO}: Microscopy for the detection, identification and quantification of malaria parasites on stained thick and thin blood films in research settings, ver 1 (2016), {World Health Organization, Geneva, Switzerland}

\bibitem{whoHelminths2021}
{WHO}: Diagnostic target product profiles for monitoring, evaluation and surveillance of schistosomiasis control programmes (2021), {World Health Organization, Geneva, Switzerland}

\bibitem{obare}
{WWARN}: Obare method calculator  ({2023}), \url{https://www.wwarn.org/obare-method-calculator}

\end{thebibliography}
%

\appendix
\section{Appendix}

\subsection{Derivation of quantitation error}
\label{appendixDeriveQuantError}
Following \cite{mehanian}, suppose our formula for estimated parasitemia, for a sample with true parasitemia $p / cV$, is:
\begin{equation}
\hat{p} = \frac{(tp + fp) - \mu(\textit{\textbf{F}}) \frac{V_E}{cV}}{\mu(\textit{\textbf{S}})} ~ \frac{cV}{V_E} 
\label{eqnQuantFormulaApp}
\end{equation}
where $tp + fp$ = the number of suspect parasites in examined volume $V$ (both true parasites and misclassified distractors), \\ 
$V_E$ is the estimate of $V$ (e.g. found by counting WBCs or RBCs),\\
and other terms are as defined previously.\\
This formula says: we count the suspects then subtract the number of FPs which we expect in $V_E$, to leave our estimate of detected parasites; we divide this by our estimate of sensitivity ${\mu(\textit{\textbf{S}})}$ to get an estimate of the actual number of parasites that were present; we then normalize by estimated volume to get $\hat{p} / cV$.

This estimate contains three sources of error relative to true parasitemia $p$: classification (of both parasites and distractors), volume estimation, and Poisson variability.
Each of these varies by sample (a particular patient's blood film). 
We will substitute these errors into Equation~\ref{eqnQuantFormula}:

Let $P_V = $ the true number of parasites in $V$, so $P_V$ has a Poisson distribution with mean $p \frac{V}{cV}$.\\
In what follows, $\Delta $ denotes the deviation of the sample's variable from the mean population value, i.e.
$S = \mu(\textit{\textbf{S}}) + \Delta S, ~F = \mu(\textit{\textbf{F}}) + \Delta S, ~V_E = V + \Delta V$, and\\
$P_V = p \frac{V}{cV} + \Delta \mathcal{P}(p,V)$ where $\mathcal{P}(p,V)$ denotes Poisson variability. 
So\\
~~~~~$tp = P_V~ (\mu(\textit{\textbf{S}}) + \Delta S) = ( p\frac{V}{cV} + \Delta \mathcal{P}(p,V)  )(\mu(\textit{\textbf{S}}) + \Delta S )$, and \\
~~~~~$fp = (\mu(\textit{\textbf{F}}) + \Delta F) \frac{V}{cV}$.
Then the quantitation error for the sample is:
\begin{equation}
\begin{multlined}
\Delta p = \hat{p} - p = -p + \\
\left[ \left( p\frac{V}{cV} + \Delta \mathcal{P}(p,V) \right)(\mu(\textit{\textbf{S}}) + \Delta S ) +
(\mu(\textit{\textbf{F}}) + \Delta F ) \frac{V}{cV} - \mu(\textit{\textbf{F}}) \frac{V_E}{cV} \right] \frac{1}{\mu(\textit{\textbf{S}})} \frac{cV}{V_E}
\end{multlined}
\label{eqnSubInErrorTermsApp}
\end{equation}
Distribute everything:\\
\begin{equation}
\begin{multlined}
\Delta p = -p + \left( p\frac{V}{cV} \frac{\mu(\textit{\textbf{S}})}{\mu(\textit{\textbf{S}})} \frac{cV}{V_E} \right) + 
\left( \Delta \mathcal{P}(p,V) \frac{\mu(\textit{\textbf{S}})}{\mu(\textit{\textbf{S}})} \frac{cV}{V_E} \right) + ~~~~~~~~~\\
~~~\left( p \frac{V}{cV}  \frac{\Delta S}{\mu(\textit{\textbf{S}})} \frac{cV}{V_E} \right) +  
\left(  \Delta \mathcal{P}(p,V)  \frac{\Delta S}{\mu(\textit{\textbf{S}})} \frac{cV}{V_E} \right) + \\
 \left( \frac{\mu(\textit{\textbf{F}})}{\mu(\textit{\textbf{S}})} \frac{V}{cV}  \frac{cV}{V_E} \right) + 
\left( \frac{\Delta F}{\mu(\textit{\textbf{S}})} \frac{V}{cV}  \frac{cV}{V_E} \right) - 
\frac{\mu(\textit{\textbf{F}})}{\mu(\textit{\textbf{S}})} \frac{V_E}{cV}  \frac{cV}{V_E} ~~~~~
\end{multlined}
\label{eqnDistributeApp}
\end{equation}
Cancel terms, and substitute $V = V_E - \Delta V$:
\begin{equation}
\begin{multlined}
\Delta p = -p + \left( p\frac{V_E - \Delta V}{V_E}   \right) + 
\left( \Delta \mathcal{P}(p,V) \frac{cV}{V_E} \right) + \\
~~~~\left( p \frac{V_E - \Delta V}{V_E}  \frac{\Delta S}{\mu(\textit{\textbf{S}})} \right) +  
\left(  \Delta \mathcal{P}(p,V)  \frac{\Delta S}{\mu(\textit{\textbf{S}})} \frac{cV}{V_E} \right) + \\
 ~~~~~  \left( \frac{\mu(\textit{\textbf{F}})}{\mu(\textit{\textbf{S}})} \frac{V_E - \Delta V}{V_E} \right) + 
\left(  \frac{\Delta F}{\mu(\textit{\textbf{S}})}  \frac{V_E - \Delta V}{V_E} \right) - 
\frac{\mu(\textit{\textbf{F}})}{\mu(\textit{\textbf{S}})} 
\end{multlined} 
\label{eqnCancelAndSwitchVApp}
\end{equation}
\begin{equation}
\begin{multlined}
\Delta{p} = -p + p - p\frac{\Delta V}{V_E}  +
\Delta \mathcal{P}(p,V) \frac{cV}{V_E} +  ~~~~~~~~~~~~~~~~~~~\\
p\frac{\Delta S}{\mu(\textit{\textbf{S}})} - p \frac{\Delta V}{V_E}  \frac{\Delta S}{\mu(\textit{\textbf{S}})}  +  
 \Delta \mathcal{P}(p,V)  \frac{\Delta S}{\mu(\textit{\textbf{S}})} \frac{cV}{V_E} + \\
 \frac{\mu(\textit{\textbf{F}})}{\mu(\textit{\textbf{S}})} - \frac{\mu(\textit{\textbf{F}})}{\mu(\textit{\textbf{S}})} \frac{\Delta V}{V_E}  + 
\frac{\Delta F}{\mu(\textit{\textbf{S}})} - \frac{\Delta F}{\mu(\textit{\textbf{S}})} \frac{\Delta V}{V_E}  - 
\frac{\mu(\textit{\textbf{F}})}{\mu(\textit{\textbf{S}})} ~~~~
\end{multlined}
\label{eqnDeltaQuant1App}
\end{equation}

To assess standard deviation of the quantitation error, $\sigma (\Delta \hat{p})$, over the population, we convert the $\Delta$ terms to std devs $\sigma(~)$. 

In the interaction terms, $\Delta V, \Delta \mathcal{P}, \Delta S$, and  $\Delta F$ are independent (relative to the other interaction term) random variables with zero mean ($\Delta \mathcal{P}$ has very close to zero mean at parasitemias relevant for quantitation vs. at LoD). 
So 

$\sigma(\Delta V \Delta S) = \sigma(\Delta V) \sigma(\Delta S)$, etc. 

As defined earlier, $\sigma(\Delta S) = \sigma(\textit{\textbf{S}})$ and $\sigma(\Delta S) = \sigma(\textit{\textbf{S}})$. 
We denote the vector of $\Delta V$s over the population as $\textit{\textbf{V}}_E$ and the vector of parasitemia estimate errors as $\textit{\textbf{p}}_E$. 

To facilitate computation later, we make an approximation when converting to std devs at population level: 
\begin{equation}
\frac{\sigma(\Delta \textit{\textbf{V}})}{V_E} \approx\frac{\sigma(\Delta \textit{\textbf{V}})}{V}
\label{eqnApproxOfVApp}
\end{equation} 
if the volume estimator (e.g. WBC or RBC counter) is decent (e.g. if standard error = 0.1 they differ by a factor $< 1.05$).  
\begin{equation}
\begin{multlined}
\sigma(\textit{\textbf{p}}_E) = 
p\frac{\sigma(\textit{\textbf{V}}_E)}{V}   ~+~ 
\sigma(\mathcal{P}(p,V)) \frac{cV}{V}  ~+~ 
p\frac{\sigma(\textit{\textbf{S}})}{\mu(\textit{\textbf{S}})}  ~+~
p \frac{\sigma(\textit{\textbf{V}}_E)}{V}  \frac{\sigma(\textit{\textbf{S}})}{\mu(\textit{\textbf{S}})}   ~+~  ~~ \\
\sigma(\mathcal{P}(p,V))  \frac{\sigma(\textit{\textbf{S}})}{\mu(\textit{\textbf{S}})} \frac{cV}{V}  ~+~  
\frac{\mu(\textit{\textbf{F}})}{\mu(\textit{\textbf{S}})} \frac{\sigma(\textit{\textbf{V}}_E)}{V}  ~+~
\frac{\sigma(\textit{\textbf{F}})}{\mu (\textit{\textbf{S}})}  ~+~ 
\frac{\sigma(\textit{\textbf{F}})}{\mu(\textit{\textbf{S}})} \frac{\sigma(\textit{\textbf{V}}_E)}{V} 
\end{multlined}
\label{eqnSigmaQuantErrorApp}
\end{equation}

Finally, note that $\frac{\sigma(\textit{\textbf{V}}_E)}{V}$ is most likely constant at usable $V$: it is the standard error of the volume estimator, and the error in the count of WBCs or RBCs will likely scale with the total true number (e.g. if $\sigma(\textit{\textbf{V}}_E) = 20 \text{ for } V = 500$, then $\sigma(\textit{\textbf{V}}_E) = 40 \text{ for } V = 1000$). 
Letting $\frac{\sigma(\textit{\textbf{V}}_E)}{V} = V_{SE}$ we have the cleaner formula
\begin{equation}
\begin{multlined}
\sigma(\textit{\textbf{p}}_E) = p V_{SE}   ~+~ 
\sigma(\mathcal{P}(p,V)) \frac{cV}{V}  ~+~ 
p\frac{\sigma(\textit{\textbf{S}})}{\mu(\textit{\textbf{S}})} ~+~ 
p V_{SE}  \frac{\sigma(\textit{\textbf{S}})}{\mu(\textit{\textbf{S}})}  ~+~  \\
\sigma(\mathcal{P}(p,V))  \frac{\sigma(\textit{\textbf{S}})}{\mu(\textit{\textbf{S}})} \frac{cV}{V}  ~+~  
\frac{\mu(\textit{\textbf{F}})}{\mu(\textit{\textbf{S}})} V_{SE}  ~+~ \frac{\sigma(\textit{\textbf{F}})}{\mu (\textit{\textbf{S}})}  ~+~ 
\frac{\sigma(\textit{\textbf{F}})}{\mu(\textit{\textbf{S}})} V_{SE} 
\end{multlined}
\label{eqnSigmaQuantErrorAppWithVse}
\end{equation}
Then the standard error of quantitation is
\begin{equation}
\begin{multlined}
\frac{\sigma(\textit{\textbf{p}}_E)}{p} = 
V_{SE}   ~+~ 
\frac{\sigma(\mathcal{P}(p,V))}{p} \frac{cV}{V}  ~+~
\frac{\sigma(\textit{\textbf{S}})}{\mu(\textit{\textbf{S}})}  ~+~ 
V_{SE}  \frac{\sigma(\textit{\textbf{S}})}{\mu(\textit{\textbf{S}})}   ~+~ ~~~~~~~~~ \\
~~~~~~~~~~~~ \frac{\sigma(\mathcal{P}(p,V))}{p} \frac{cV}{V} \frac{\sigma(\textit{\textbf{S}})}{\mu(\textit{\textbf{S}})}  ~+~  
\frac{V_{SE}}{p}\frac{\mu(\textit{\textbf{F}})}{\mu(\textit{\textbf{S}})}  ~+~ \frac{1}{p}\frac{\sigma(\textit{\textbf{F}})}{\mu (\textit{\textbf{S}})}  ~+~ 
\frac{V_{SE}}{p}\frac{\sigma(\textit{\textbf{F}})}{\mu(\textit{\textbf{S}})} \\
\end{multlined}
\label{eqnSigmaQuantErrorAppWithVse}
\end{equation}
Grouping and reordering terms gives
\begin{equation}
\begin{multlined}
\frac{\sigma(\textit{\textbf{p}}_E)}{p} = 
V_{SE}   ~+~ 
\frac{\sigma(\textit{\textbf{S}})}{\mu(\textit{\textbf{S}})} \left(1 + V_{SE}\right)  ~+~ ~~~~~~~~~~~~~~~~~~~\\
~~~~~~~~~\left(1 + \frac{\sigma(\textit{\textbf{S}})}{\mu(\textit{\textbf{S}})} \right)
 \sqrt{\frac{1}{p}\frac{cV}{V}}   ~+~  
\frac{V_{SE}}{p}\frac{\mu(\textit{\textbf{F}})}{\mu(\textit{\textbf{S}})}   ~+~
\frac{\sigma(\textit{\textbf{F}})}{\mu(\textit{\textbf{S}})} \frac{(1 + V_{SE})}{p}
\end{multlined}
\label{eqnSigmaQuantErrorFinal}
\end{equation}
where we used the fact that for a Poisson distribution the variance equals the mean, so $\sigma(\mathcal{P}(p,V)) = \sqrt{p\frac{V}{cV}}$.

This is Equation~\ref{eqnStdQuantError} in the main text. 

This equation has the following structure: (i) the first 2 terms involve only sensitivity and volume estimation, and are constant for all $\{p,V\}$; (ii) only the third term involves Poisson variability, and it decreases with $\sqrt{pV}$; and (iii) only the last 2 terms involve FP rates, and they decrease with $p$.

If Poisson and volume estimation error are ignored and only classification error is considered, i.e. when $\sigma(\mathcal{P}) = V_{SE} = 0$, then most terms disappear and it becomes the formula for the standard error of quantitation due to classifier inaccuracy as derived in \cite{thinGhtc2019SuppInfo}:
\begin{equation}
\frac{\sigma(\textit{\textbf{p}}_E)}{p} = \frac{\sigma(\textit{\textbf{S}})}{\mu (\textit{\textbf{S}})} + \frac{\sigma(\textit{\textbf{F}})}{\mu (\textit{\textbf{S}})}\frac{1}{p}
\label{eqnStdErrorModel}
\end{equation}  
%
%
~  \\ 
\subsection{Errors in estimating examined volume}
\label{secEstimateVolumeApp}

As noted in \cite{frontierMetrics}, another source of quantitation error is error in estimating the volume examined.
According to WHO protocols, on thick films blood volume is estimated by counting WBCs and using the approximation 8000 WBCs/$\mu L$ \cite{whoMicroscopy2016} (6000/$\mu L$ in Peru \cite{peruMalaria}).
It is estimated on thin films by examining microscope fields-of-view, ballparking RBC count per field-of-view, then using the approximation 5e6 RBCs/$\mu L$ \cite{whoMicroscopyQuantSOP}. 
In research situations a grid system is sometimes used, where 5 $\mu L$ of blood is evenly spread across a grid of fixed size, and volume is estimated by area \cite{whoMalariaResearch2016}.

Two details of protocol acknowledge an imprecision in human quantitations for malaria.
These details may reflect an expectation of errors in parasite counting and volume estimation in addition to known Poisson error:
First, when possible two or more manual quantitations are averaged \cite{obare}. 
This carries high operational cost however, and is not typical for diagnostic settings.
Second, the WHO proficiency standards define someone whose quantitations are within 25\% of ground truth at least half the time on a defined set of 15 blood films as having ``Level 1'' proficiency \cite{whoQualityAssuranceV2}.

In the WBC/RBC counting cases, machines have a substantial advantage. 
Although expert humans are extremely skilled, one can reasonably expect non-trivial error when manually counting hundreds of cells while moving through microscope fields of view and concurrently tallying parasite counts, or when depending on bulk estimates of RBCs per field without careful counting.
We are not aware of any studies of human counting error in this context, so we left it out of our analysis above.
However, perfect human volume estimation is likely an unrealistic assumption.
%
%
\subsection{Code to plot Poisson distributions}
\label{appendixCode} 

import numpy as np  \\
from matplotlib import pyplot as plt  \\
from scipy.stats import poisson  \\\\
pPerUL = 100  \\
vols = np.array((0.01, 0.02, 0.05, 0.1))  \\
numWbcs = vols * 8000  \\
numRbcs = vols * 5000  \\
mu = pPerUL * vols  \\
numDraws = 10000  \\
k = np.arange(0, 20)  \\\\
probK = np.zeros((len(k), len(vols)))  \\
cumProbK = np.zeros((len(k), len(vols)))  \\
stdProbK = np.zeros(len(vols))  \\\\
for i in range(len(vols)):  \\
    for j in range(len(k)):  \\
        probK[j, i] = poisson.pmf(k = k[j], mu=mu[i])  \\
        cumProbK[j,i] = poisson.cdf(k = k[j], mu=mu[i])  \\
        stdProbK[i] = np.std(poisson.rvs(mu = mu[i],size=numDraws ))  \\\\
stdError = stdProbK / mu  \\\\
\# plot distributions:  \\
print('vols      = ' + str(vols))  \\
print('std error = ' + str(np.round(stdError,2)))  \\\\
tickKwargs = {'fontweight':'bold','fontsize':12}    \\
legendKwargs = {'fontsize':12}  \\\\
plt.figure()   \\
plt.xlabel('\# of parasites in examined volume (true P = 100 p/uL)',  \\
           fontweight='bold', fontsize=12)  \\
plt.ylabel('Probability', fontweight='bold', fontsize=12)  \\
for i in range(len(vols)):  \\
    plt.plot(k, probK[:, i], linewidth=2,  \\
             label = str(vols[i]) + ' uL (' + str(int(numWbcs[i])) + ' wbcs or ' + \  \\
                 str(int(numRbcs[i])) + 'k rbcs)')  \\
plt.legend(**legendKwargs)  \\
plt.xticks(range(0,21,2), **tickKwargs)  \\
plt.yticks(np.arange(0, 0.5, 0.1), **tickKwargs)  


\subsection{Code to plot standard error of quantitation}
\label{appendixStdErrorQuantCode} 

import os  \\
import numpy as np  \\
from matplotlib import pyplot as plt, rc  \\  \\
\# parasitemias:  \\
p = list(range(100, 1001, 50)) + list(range(1000, 10000, 500)) + \\ 
    list(range(10000, 150000, 2000))  \\
p = np.array(p)  \\
\# Volumes examined:  \\
Vrbc = 0.0004   \# 2000 RBCs if > 16k p/uL.  TDR  \\
Vwbc = 0.0625    \# 500 WBCs if p < 16k/uL.  TDR.  \\
VwbcWho = 0.025    \# 200 WBCs if p > 400, < 16k  \\
Vauto = 0.125   \# 1000 WBCs  \\
Vexam = np.array([0.0625, 0.1, 0.25, 0.5]) \# volume examined by algorithm  \\  \\
\# Algorithm performance statistics:  \\
muS = 0.95  \\
sigmaS = 0.03  \\
muF = 50  \\
sigmaF = 10  \\
Vse = 0.02   \# ie std dev = 2\% of total count  \\
VseHuman = 0.02  \\  \\
\# Various fixed Vs:  \\
PoisVrbc = np.sqrt(1 / (p * Vrbc))  \\
PoisVwbc = np.sqrt(1 / (p * Vwbc))  \\
PoisVauto = np.sqrt(1 / (p * Vauto))  \\  \\
\# Populate for range of Vs:  \\
PoisVexam = np.zeros((len(Vexam), len(p)))  \\
PoisAndSigSTerm = np.zeros((len(Vexam), len(p)))  \\
muFTerm = np.zeros((len(Vexam), len(p)))  \\
sigmaFTerm = np.zeros((len(Vexam), len(p)))  \\  \\
for i in range(len(Vexam)):  \\
    PoisVexam[i, :] = np.sqrt(1 / (p * Vexam[i]))  \\
    PoisAndSigSTerm[i, :] = (sigmaS / muS) * np.sqrt(1 / (p * Vexam[i]))  \\
    muFTerm[i, :] = (Vse / p) * (muF / muS)  \\
    sigmaFTerm[i, :] = sigmaF / muS *(1 + Vse) / p  \\
\# Constant:  \\
sigmaSConstantTerm = (sigmaS / muS) * (1 + Vse)  \\  \\
totalStdError = np.zeros((len(Vexam), len(p)))  \\
for i in range(len(Vexam)):  \\
    totalStdError[i, :] = Vse + sigmaSConstantTerm + \  \\
        PoisVexam[i,:] + PoisAndSigSTerm[i, :] + muFTerm[i, :] + sigmaFTerm[i, :]  \\  \\
\#\%\% Plot Poisson error (only) for select V's including protocols:  \\
plt.figure()  \\
rc('font',weight='bold')  \\
\# Vwbc:  \\
inds = np.where(p < 16001)[0]  \\
plt.semilogx(p[inds], PoisVwbc[inds], color='r', label='Protocol, thick, 500 WBCs')  \\
\# Vrbc:  \\
inds = np.where(p >= 16000)[0]  \\
plt.semilogx(p[inds], PoisVrbc[inds], linestyle='--', color='r',  \\
             label='Protocol, thin, 2000 RBCs')  \\
\# Autoscope:  \\
plt.semilogx(p, PoisVauto, 'b', label='Machine, 1000 WBCs')  \\  \\
plt.legend()  \\
plt.xlabel('parasites / uL', fontweight='bold', fontsize=12)  \\
plt.ylabel('std error', fontweight='bold', fontsize=12)  \\  \\
\#\%\% Plot some of the important terms for one Vexam:  \\
ind = np.where(Vexam == 0.1)[0][0]  \\
plt.figure()  \\
rc('font', weight='bold')  \\  \\
plt.semilogx(p, sigmaSConstantTerm *np.ones(len(p)),'m',  \\
             label='sigmaS / muS term')  \\
plt.semilogx(p, Vse *np.ones(len(p)),'k', label='Vse term')  \\
plt.semilogx(p,PoisAndSigSTerm[ind,:],'b', label='Poisson * (sigmaS / muS) term')  \\
plt.semilogx(p,PoisVexam[ind,:],'c', label='Poisson only term')  \\
plt.semilogx(p,sigmaFTerm[ind,:],'g', label='sigmaF / muS term')  \\
plt.semilogx(p,muFTerm[ind,:],'r', label='muF / muS term')  \\  \\
plt.legend()  \\
plt.xlabel('parasites / uL', fontweight='bold', fontsize=12)  \\
plt.ylabel('std error', fontweight='bold', fontsize=12)  \\
plt.title('Components of std error equation for: ' +  
          'V = ' + str(Vexam[ind]) + ', Vse = ' + str(Vse) + \\ 
          'muS = ' + str(muS) + ', sigmaS = ' + str(sigmaS)  +  
          ', muF = ' + str(muF) + ', sigmaF = ' + str(sigmaF),  
          fontweight='bold')  \\  \\
\#\%\% Plot Poisson error for protocol and total error for selection of Vs:  \\
plt.figure()  \\
rc('font',weight='bold')  \\  \\
\# Vwbc:  \\
inds = np.where(p < 16001)[0]  \\
plt.semilogx(p[inds], PoisVwbc[inds], color='r',  \\
             label='Protocol, thick, ' + str(Vwbc) + ' uL')  \\
inds = np.where(p < 16001)[0]  \\
\# Vrbc:  \\
inds = np.where(p >= 16000)[0]  \\
plt.semilogx(p[inds], PoisVrbc[inds], linestyle=':', color='r',  \\
             label='Protocol, thin, ' + str(Vrbc) + ' uL')  \\
\# Vwbc + some volume estimation error:  \\
inds = np.where(p < 16001)[0]  \\
plt.semilogx(p[inds], PoisVwbc[inds] + VseHuman, color='r', linestyle='--',  \\
              label='Protocol, thick, ' + str(Vwbc) + ' uL plus se')  \\
\# Machine  \\
for i in range(len(PoisVexam)):  \\
    plt.semilogx(p, totalStdError[i], 'b', label=str(Vexam[i]) + ' uL')  \\
\# Rerun Vwbc to make them foreground:  \\
inds = np.where(p < 16001)[0]  \\
plt.semilogx(p[inds], PoisVwbc[inds], color='r')  \\
plt.semilogx(p[inds], PoisVwbc[inds] + VseHuman, color='r', linestyle='--')  \\
plt.legend()  \\
plt.xlabel('parasites / uL', fontweight='bold', fontsize=12)  \\
plt.ylabel('std error', fontweight='bold', fontsize=12)

\end{document}